\documentclass[10pt,twocolumn,letterpaper]{article}

\usepackage[pagenumbers]{cvpr} 

\usepackage[dvipsnames]{xcolor}

\usepackage{graphicx}
\usepackage{amsmath}
\usepackage{amssymb}
\usepackage{booktabs}
\usepackage{makecell}
\graphicspath{{./figures/}}

\definecolor{cvprblue}{rgb}{0.21,0.49,0.74}
\usepackage[pagebackref,breaklinks,colorlinks,citecolor=cvprblue]{hyperref}

\def\paperID{*****} 
\def\confName{CVPR}
\def\confYear{2024}

\title{LangOcc: Self-Supervised Open Vocabulary Occupancy Estimation via Volume Rendering}

\author{Simon Boeder\\
Robert Bosch GmbH\\
{\tt\small simon.boeder@de.bosch.com}
\and
Fabian Gigengack\\
Robert Bosch GmbH\\
{\tt\small fabian.gigengack@de.bosch.com}
\and
Benjamin Risse\\
University of M\"unster\\
{\tt\small b.risse@uni-muenster.de}
}

\begin{document}
\maketitle

\begin{abstract}

The 3D occupancy estimation task has become an important challenge in the area of vision-based autonomous driving recently.
However, most existing camera-based methods rely on costly 3D voxel labels or LiDAR scans for training, limiting their practicality and scalability.
Moreover, most methods are tied to a predefined set of classes which they can detect.
In this work we present a novel approach for open vocabulary occupancy estimation called \textit{LangOcc}, that is trained only via camera images, and can detect arbitrary semantics via vision-language alignment.
In particular, we distill the knowledge of the strong vision-language aligned encoder CLIP into a 3D occupancy model via differentiable volume rendering. 
Our model estimates vision-language aligned features in a 3D voxel grid using only images.
It is trained in a self-supervised manner by rendering our estimations back to 2D space, where ground-truth features can be computed.
This training mechanism automatically supervises the scene geometry, allowing for a straight-forward and powerful training method without any explicit geometry supervision.
LangOcc outperforms LiDAR-supervised competitors in open vocabulary occupancy by a large margin, solely relying on vision-based training.
We also achieve state-of-the-art results in self-supervised semantic occupancy estimation on the Occ3D-nuScenes dataset, despite not being limited to a specific set of categories, thus demonstrating the effectiveness of our proposed vision-language training.
\vspace{-5mm}
\end{abstract}

\section{Introduction}
\label{sec:intro}

Object detection is a fundamental task in autonomous driving, enabling vehicles to understand and navigate their surroundings.
Traditionally, these tasks have been trained on predefined sets of classes, limiting their ability to fully comprehend complex and dynamic environments.
To overcome this limitation, recent advancements have introduced 3D occupancy estimation, a popular method that represents scene geometry using a voxel grid \cite{zhang2024vision, jiang2024openocc, tian2023occ3d, chen2023end, gan2023simple}.
This approach allows for geometry-based \emph{generic} object detection, enabling autonomous vehicles to perceive any structure in their environment.
However, most existing 3D occupancy estimation methods rely on expensive 3D ground-truth labels \cite{huang2023tri, li2023voxformer, zhang2023occformer}.
This requirement poses a significant challenge, as acquiring accurate 3D labels for large-scale datasets is both resource-intensive and impractical.
Moreover, existing benchmarks usually only reflect a limited predefined set of classes.
Consequently, there is a pressing need for novel methods to efficiently train occupancy models without relying on 3D labels, for example via self-supervised learning.
While some efforts have been made to avoid voxel labels, they either still necessitate labeled LiDAR point clouds or involve complex pseudo ground-truth generation techniques \cite{pan2023renderocc,huang2023selfocc,zhang2023occnerf}.
Furthermore, despite the ability to capture any geometry, the semantic understanding of these methods remains tied to a predefined set of classes.
These limitations hinder the adaptability and flexibility of autonomous systems in comprehending diverse and evolving environments.

In this paper we propose a novel self-supervised occupancy estimation method which aligns geometric estimations with open vocabulary natural language features, hence allowing representations of any semantics and therefore eliminating the need for 2D or 3D semantic labels.
To achieve this, we leverage the power of the popular CLIP model \cite{radford2021learning} and distill its representational power into 3D space through volume rendering.
In particular, instead of predicting the probabilities of predefined classes, our model estimates vision-language aligned features per voxel.
The model is trained by rendering these features in a differentiable manner from the 3D voxel space back to the 2D image space, where they are supervised by features precomputed by the off-the-shelf vision-language encoder CLIP \cite{radford2021learning}.
The source code will be released soon.

In summary, our contributions are:
\begin{itemize}
    \item \textbf{Open vocabulary occupancy:} A novel vision-only architecture to model arbitrary geometries and semantics by aligning the semantic feature space with natural language, hence decoupling occupancy representations from predefined semantic class definitions.
    \item \textbf{Self-supervised learning:} Inspired by NeRF~\cite{mildenhall2021nerf}, LangOcc trains language features and 3D scene geometry jointly and eliminates the need for 3D ground-truth labels. 
    As a consequence, the proposed method can be trained with images only.
    Our model generalizes to estimate geometry and semantics in a \textbf{zero-shot manner}, without per-scene optimization like NeRF-approaches.
    \item \textbf{Feature subspace learning:} In addition we introduce a specialised dimensionality reduction strategy to increase segmentation performance when a set of task-specific classes is available.
    \item \textbf{State-of-the-art performance:} LangOcc outperforms competitors on open vocabulary occupancy estimation by a large margin, and achieves state-of-the-art results in self-supervised semantic occupancy estimation.
\end{itemize}

\section{Related Work}
Three different lines of work are particularly relevant for our proposed method, namely object detection, occupancy estimation and open vocabulary perception.
\subsection{Camera-based 3D Object Detection}
Vision-based 3D object detection is crucial for autonomous applications and a widely studied field.
Most recent approaches transforms extracted 2D image features from a single or multiple views into a common 3D space (e.g. a Birds-Eye-View grid; BEV) where objects boxes are estimated.
One popular approach is to lift 2D image features into 3D by estimating a depth distribution \cite{philion2020lift, huang2021bevdet, huang2022bevdet4d}, while other methods such as \cite{li2022bevformer,jiang2023polarformer} project learned 3D queries onto the image plane to sample features.
Methods like \cite{wang2023exploring, liu2022petr} do not explicitly project features to 3D but instead follow an object-centric approach.
All of these methods are typically trained to detect specific types of objects, and therefore do not provide a comprehensive understanding of the entire scene.
This limitation has prompted the exploration of the more general occupancy estimation paradigm, which aims to perceive and understand the complete geometry of the scene.

\subsection{Camera-based 3D Occupancy Estimation}
Vision-based occupancy prediction, also known as semantic scene completion, involves estimating a dense representation of the 3D scene in terms of geometry and semantics from a set of input images \cite{behley2019iccv, tian2023occ3d, wang2023openoccupancy}.
Pioneering works on 3D occupancy estimation extend the well-known concepts of object detection to 3D space, e.g., by lifting the BEV into a voxel grid \cite{huang2023tri, huang2022bevdet4d, tong2023scene, cao2022monoscene, li2023voxformer}.
Following approaches mostly focus on efficient supervision \cite{yu2023flashocc, lu2023octreeocc, liu2024fully}, label efficiency \cite{boeder2024occflownet, pan2023renderocc, gan2023simple} or performance improvements via specific model designs \cite{li2023fb, zhang2023occformer, jiang2023symphonize, tan2024geocc, Zhao_2024_CVPR}.
As the 3D occupancy prediction task is inherently more complex, most models rely on 3D ground truth data, which can be challenging and resource-intensive to obtain.
Consequently, there have been efforts to explore self-supervised learning approaches for training occupancy models using only image data \cite{huang2023selfocc, zhang2023occnerf}.
Specifically, volume rendering supervision (inspired by, e.g., NeRF \cite{mildenhall2021nerf} and classical volume rendering \cite{kajiya1984ray}) has demonstrated great potential as a training mechanism for occupancy estimation models.
It enables the simultaneous supervision of geometry and semantics using 2D labels, which are considerably easier to acquire than 3D voxel labels \cite{zhang2023occnerf, huang2023selfocc, boeder2024occflownet}.
Despite these advancements, existing methods are often constrained by a set of predefined classes or rely on pretrained models to generate ground truth, lacking a true generic scene representation.

\subsection{Open Vocabulary Perception}
The goal of open vocabulary perception (or similarly \textit{zero-shot semantic segmentation}) is to detect or segment object classes that were not explicitly seen during training, given a natural language query.
With the help of multi-modal models like CLIP \cite{radford2021learning}, many approaches have been developed in this regard.
A common method is to extend CLIP to produce pixel-level features instead of a single image wide feature.
MaskCLIP \cite{zhou2022extract} modifies the last pooling layer of CLIP, while LERF \cite{kerr2023lerf} and CLIP-FO3D \cite{zhang2023clip} extract patch-wise CLIP embeddings for an image in a sliding-window fashion.
Further, methods like \cite{li2022language,ghiasi2022scaling,liang2023open} train networks on pixel-level segmentation datasets and distill CLIP features simultaneously.
OVR \cite{zareian2021open} trains a generalizable 2D object detector with language pretraining, while ViLD \cite{gu2021open} distills CLIP knowledge into a 2-stage detector.
OWL-ViT \cite{minderer2022simple} directly attaches a detector to the CLIP image encoder.
To enable 3D open vocabulary perception, distillation of vision-language features into NeRFs \cite{kerr2023lerf} or Gaussian Splatting \cite{qin2024langsplat, zhou2024feature} have been explored, however these are only trained on a per-scene basis.
CLIP-Fo3D \cite{zhang2023clip} directly distills extracted vision-language features into a given 3D point cloud via projection.
Recently, there have also been efforts for open vocabulary occupancy estimation similar to our work.
Most notably, POP-3D \cite{vobecky2024pop} trains a model to predict 3D occupancy and 3D vision-language features given just images, but require LiDAR scans during training.
Similarly, OVO \cite{tan2023ovo} aligns voxel predictions with precomputed feature maps, but lacks geometry supervision and is only designed for small and simple scenes.
Finally, OpenOcc \cite{jiang2024openocc} also represent the scene with voxels, but perform scene reconstruction on a per-scene basis like LERF \cite{kerr2023lerf}.

\section{Methodology}

\subsection{Problem Definition}\label{sec:problem_definition}
Given a set of RGB images $I = \{I^1, I^2, ..., I^N\}$, the objective is to estimate the surrounding environment as a 3D voxel representation $V$ on a defined grid.
Each voxel in the representation is assigned an occupancy probability $V_{\sigma} \in [0, 1]^{X \times Y \times Z}$.
Additionally, a vision-language aligned feature vector is estimated for each voxel $V_{\psi} \in \mathbb{R}^{X \times Y \times Z \times L}$ to model the semantics of the scene in a generic manner.
These voxel features can be utilized in various downstream tasks, such as \textit{zero-shot semantic occupancy estimation} or \textit{open vocabulary retrieval}.

\subsection{Model Architecture}\label{sec:model_arch}
\begin{figure*}
    \centering
    \includegraphics[page=1, trim=0cm 7.01cm 6.8cm 0cm, clip, width=1\textwidth]{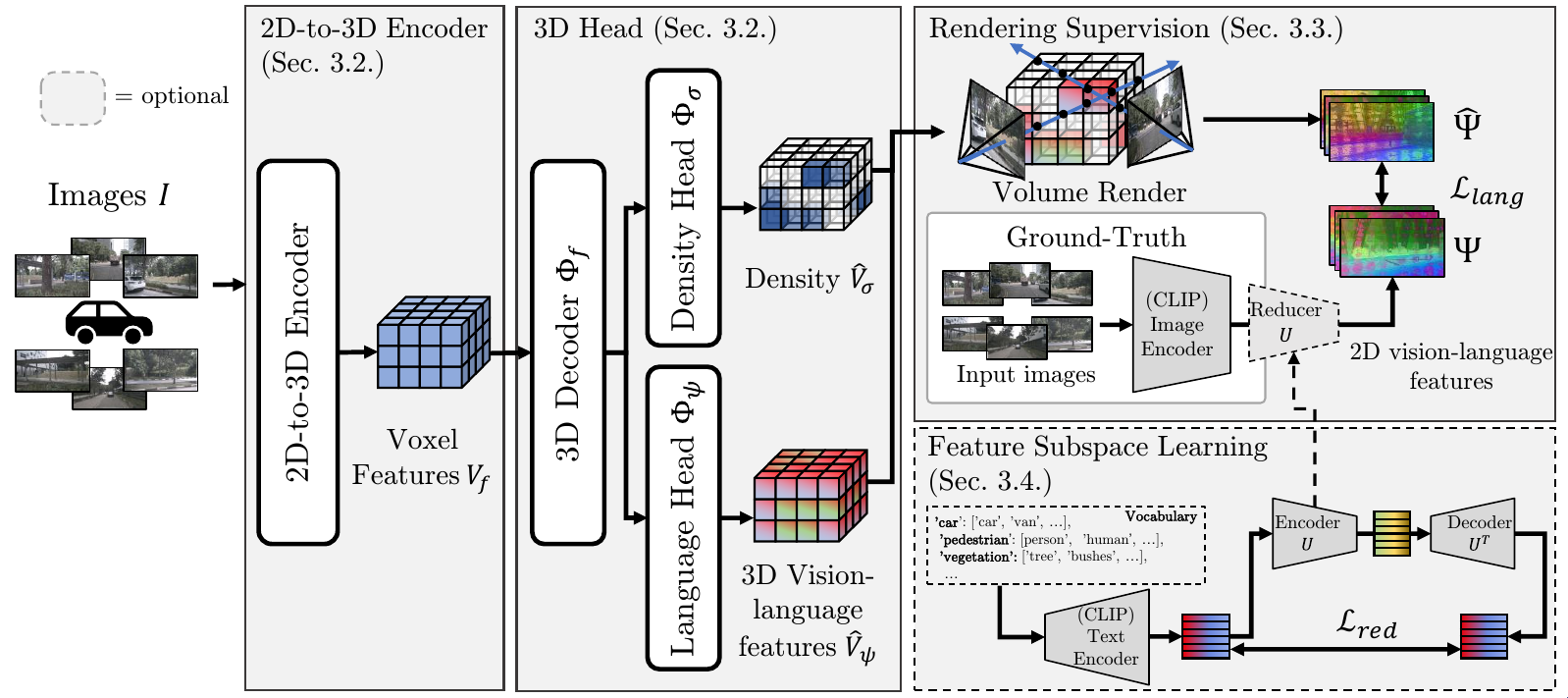}
    \caption{
        \textbf{Architecture of the proposed model.}
        A set of images is first transformed to 3D voxel features via BEVStereo \cite{li2023bevstereo} and a 3D CNN decoder.
        Next, two separate heads estimate the density probabilities and the generic scene semantics as vision-language features.
        The model is trained via differentiable volume rendering, using a loss between rendered estimated features and precomputed 2D features from MaskCLIP \cite{zhou2022extract}.
        Optionally, to increase training efficiency and performance at the cost of expressiveness, \textit{feature subspace learning} can be applied using a predefined vocabulary.
    }
    \label{fig:architecture}
    \vspace{-4mm}
\end{figure*}

The proposed model is outlined in \cref{fig:architecture}.
Initially, the input images $I$ are transformed into 3D voxel features $V_f$ using the prominent 2D-to-3D transformation network BEVStereo \cite{li2023bevstereo}, similar to previous works.
However, note that any other 2D-to-3D encoder, like \cite{li2022bevformer, huang2021bevdet, huang2023tri}, could be used instead.
Afterwards, these voxel features are used to predict the density $V_\sigma$ and the language aligned features $V_\psi$ using a 3D CNN decoder and two separate MLP heads.
The entire model is supervised using volume rendering supervision by rendering the estimated 3D features back to the 2D image space and comparing them with precomputed vision-language features (\cref{sec:render}).
Additionally, in \cref{sec:lowerspace}, an optional method to enhance detection performance and training efficiency by pretraining a dimensionality reduction encoder on a given vocabulary is presented.

\paragraph{2D-to-3D Encoder} \label{sec:2dto3d}
Image features are first extracted from the input images $I$ using a pretrained backbone architecture.
Next, the features of the current frame and a specified amount of previous frames, used for temporal propagation, are projected into 3D space using the known camera parameters and depth estimation. 
The 3D features are then pooled to a common 3D voxel grid of features $V_f \in \mathbb{R}^{X \times Y \times Z \times C}$, where $X,Y,Z$ represent the resolution of the grid and $C$ denotes the size of the latent dimension.
This architecture is based on BEVStereo \cite{li2023bevstereo}, except that the features are pooled into a 3D voxel grid instead of a 2D Birds-Eye-View grid.

\paragraph{3D Head} \label{sec:3dheads}
The voxel features $V_f$ are processed by a 3D CNN decoder $\Phi_f$, which computes local interactions to refine the features.
Subsequently, for each voxel two separate MLP heads $\Phi_\sigma$ and $\Phi_\psi$ calculate the density probability $\sigma$ and a vision-language feature $\psi \in \mathbb{R}^{L}$, where $L$ represents the feature dimension size.
The outputs of $\Phi_\sigma$ are transformed to probabilities using the sigmoid function denoted as $s(\cdot)$
Essentially, the scene geometry is represented by the density probabilities $V_\sigma$, which can also be interpreted as occupancy probabilities, while the semantics of the scene is represented by the vision-language features $V_\phi$, which is by design no tied to any specific set of classes.

\begin{align}
  &V_\sigma = s ( \Phi_\sigma(\Phi_f(V_f))) \in [0, 1]^{X \times Y \times Z} \; \\
  &V_\psi = \Phi_\psi( \Phi_f(V_f)) \in \mathbb{R}^{X \times Y \times Z \times L} \;
  \label{eq:decoder}
\end{align}

As will be explained in \cref{sec:render}, this separation is required to enable training via volume rendering, which automatically supervises geometry without any explicit loss.

\subsection{Volume Rendering Supervision}\label{sec:render}

To supervise the entire model, we use differentiable volume rendering, a technique that gained popularity with the introduction of NeRF \cite{mildenhall2021nerf}.
Similar to recent works \cite{pan2023renderocc, boeder2024occflownet, huang2023selfocc, zhang2023occnerf}, instead of overfitting a network on a single scene, we use volume rendering as a differentiable operation to bring our predictions from the 3D voxel space back to the 2D image space, where ground truth labels are much easier to acquire.

After estimating the volumes $V_\sigma$ and $V_\psi$, for each camera $i$ in the current frame, a set of 3D rays $D^i_r$ is generated, each originating from the camera origin $o_i$ in the direction $d_i(u,v)$ of a pixel $(u, v)$ of the image into the 3D voxel grid, using the camera extrinsic and intrinsic parameters.
For each ray $r$, we then sample a number of 3D points $r(t) = o + td$ at different distances $t$ along the ray and collect the density probabilities $\sigma(r(t))$ and language features $\psi(r(t))$ at these points from the predicted volumes using trilinear interpolation.
We then accumulate the language features along each ray to render them to a single feature using the traditional differentiable rendering formulation \cite{kajiya1984ray}, as in NeRF \cite{zhang2023occnerf}.
Specifically, a rendering weight $w(r(t))$ is computed for each sampled point on the ray by accumulating the interpolated density:

\begin{align}
	&w(r(t)) = T\left(r(t)\right) \left(1 - \exp(- \sigma(r(t)) \delta_t) \right) \textrm{, with} \\
	&T(r(t)) = \exp\left(-\sum_{j=1}^{t-1} \sigma(r(j)) \delta_j \right),
\end{align}

where $T(r(t))$ represents the cumulative transmittance along the ray up to $t$ and $\delta_t$ is the distance between the current and next sample. 
This weight determines the contribution of each point to the final value based on their estimated density.
Given this weight, the final rendered 2D vision-language features can be computed by summing up the point features multiplied by their rendering weight.
\begin{align}
    \hat{\Psi}(r) = \sum_{t=1}^N w(r(t)) \psi(r(t)) 
\end{align}

\paragraph{Loss Function}
After rendering the 3D features $V_\psi$ into 2D features $\hat{\Psi}$, a loss can be computed between the estimated features and some 2D ground truth features that we extract from the same input image using a vision-language aligned image encoder, such as CLIP \cite{radford2021learning}.
In this work, we adopt the method proposed in \cite{vobecky2024pop} and extract pixel-level CLIP features using MaskCLIP \cite{zhou2022extract}.
For each ray, we fetch the target feature $\Psi(r)$ via bi-linear interpolation of the precomputed MaskCLIP feature map $I^i_\psi$ at the pixel coordinate $(u,v)$.
As a loss function, we propose the \textit{Cosine Similarity Guided MSE}, which is a combination of the cosine similarity loss and the mean-squared error loss function.
We have found that the MSE loss function has a much easier-to-optimize loss landscape, while the cosine similarity gives a better notion of how close the embeddings are in the CLIP space.
Therefore, we optimize the MSE loss weighted by the cosine distance $\mathcal{C}$ for each ray, so that features already estimated well have less influence, while features with low cosine similarity to the target have a higher influence on the final loss:

\begin{align}
    &\mathcal{L}_{lang}(\hat{\Psi}, \Psi) = \mathcal{C}(\hat{\Psi}, \Psi) * ||\Psi - \hat{\Psi}||^2 \\
    &\mathcal{C}(\hat{\Psi},\Psi) = (1 - \frac{\hat{\Psi} * \Psi}{||\hat{\Psi}||_2* ||\Psi||_2})
\end{align}
Note that in the implementation of this loss, we put a \textit{stop\_grad} on the cosine distance, such that it is not used during backpropagation.
\textbf{Essentially, we distill the knowledge of a strong pretrained 2D vision-language encoder into a 3D voxel-based model via volume rendering, while maintaining the language alignment functionalities.}

Simultaneously, the model is forced to learn correct scene geometry estimations in order to be able to render the features from 3D into different cameras.
Therefore, the scene geometry estimation is learned automatically, without any additional loss.
It is important to note that the volume rendering technique is only applied during training.
During inference, the model just takes the 2D images as input and outputs the scene geometry and 3D vision-language features.

\paragraph{Temporal Rendering}\label{sec:temporal_render}
In order to estimate the 3D geometry of the scene correctly, the volume rendering supervision method introduced above requires that the voxels are seen from multiple rays, as the depth of a ray is otherwise ambiguous.
However, the field of view overlap between different cameras in a multi-view setup is usually very low (e.g., nuScenes \cite{caesar2020nuscenes}), which aggravates the learning of the correct densities without any explicit geometry supervision.
To address this, we adopt the temporal rendering approach of recent works \cite{pan2023renderocc,boeder2024occflownet,huang2023selfocc} and additionally render 2D feature maps for a set of temporally adjacent input images $I_t = \{I_{-t}, ..., I_{-1}, I_{+1}, ..., I_{+t}\}$ during training.
For each frame during training, we also generate rays for all temporal frames in a predefined time horizon, and compute the same loss as described above. 
As we show in the experiments, this temporal rendering approach is crucial for the model to simultaneously learn geometry and semantics from just the feature distillation loss.
However, rendering temporally adjacent frames introduces errors due to dynamic objects.
We always render our predictions for the current time step, but compute a loss to ground truth feature maps from adjacent time steps, where objects might have moved.
As previous work has shown, compensating these errors can lead to better performance \cite{boeder2024occflownet}, but requires either ground truth flow data or an additional training task.
As this affects only a fraction of voxels, we accept the false supervisory signals from temporal inconsistencies in this work and leave this problem for future work.

\subsection{Feature Subspace Learning} \label{sec:lowerspace}
While vision-language features offer strong representational power for scene semantics, training a model with the high-dimensional embedding space of vision-language encoders like CLIP imposes a significant computational and memory overhead.
Also, not every task necessitates the full expressiveness of the vision-language encoder.
In some cases, the requirement may be to detect a specific set of object categories (e.g. the zero-shot occupancy estimation task).
Therefore, we propose a method adopted from \cite{kobs2023indirect} and train an autoencoder to reduce the embedding space of CLIP to a smaller, task-specific subspace.
This offers a trade-off between open vocabulary expressiveness of the full embedding space and a more efficient and specialised lower dimensional space.
A lower dimensional subspace specifically modeled for the task at hand can also increase segmentation performance as well as training speed.

Prior to training of our proposed model, we train a single linear transformation $U \in \mathbb{R}^{L \times L'}$ that maps from the original feature space $L$ to the lower dimensional space $L'$ as the encoder, and use the transposed transformation $U^T$ as the decoder. 
Thus, the encoder and decoder share weights, which forces the matrix $U$ to become orthogonal and reduces the overall amount of parameters to prevent overfitting.
This autoencoder is trained solely on the vision-language features $t_i \in \mathbb{R}^{L} \text{ for } i \in \{1, ..., n\}$ of a set of $n$ text prompts from a predefined vocabulary, computed via the corresponding text encoder of the vision-language encoder.
The same loss as in \cite{kobs2023indirect} is used to train $U$.

\begin{align}
    &t_i' = \frac{t_i U}{||t_i U ||} \quad \quad \hat{t}_i = \frac{t_i' U^T}{||t_i' U^T ||} \\
    & \mathcal{L}_{red} = \frac{1}{n} \sum_{i=1}^n \text{arccos} (t_i, \hat{t}_i).
\end{align}

The dataset consists of just a few text prompts, enabling the training of $U$ within seconds.
By defining a vocabulary before training, we ensure that the lower-dimensional subspace $\bar{L}$ can focus on the required information and does not model unnecessary features.
We can freely define the classes to be detected before the training.
Furthermore, we are not bound to either ground truth classes \cite{caesar2020nuscenes, tian2023occ3d, tong2023scene, wang2023openoccupancy} or pretrained object detectors \cite{huang2023selfocc, zhang2023occnerf}.
After the autoencoder is trained, we can use the encoder $U$ to reduce the dimensionality of the ground truth vision-language features $L$ of the images, as the text and vision features are inherently aligned.
We then reduce the dimensionality of our language head accordingly and train the model as before.
This method thus offers a much more efficient training when detecting certain classes is required, by simply defining a vocabulary of categories of interest, without any overhead.
We also refer to the trained encoder $U$ as the \textit{reducer} model.

\subsection{Inference}\label{sec:inference}
At inference, the estimated embeddings can be used in a versatile way.
In this work, we solve the tasks of \textit{3D open vocabulary retrieval} and \textit{zero-shot semantic occupancy estimation}.
Results are provided in \cref{sec:experiments}.

\paragraph{3D Open Vocabulary Retrieval}
We compute the language feature of a given text query using the text encoder, and then compute the similarity of this query feature with each voxel embedding via the dot product.
The resulting similarities can be visualized (e.g., by using a heatmap), or used for binary classification using a threshold.

\paragraph{Zero-shot Semantic Occupancy Estimation}
Similarly, we can assign each voxel a category by defining a vocabulary that consists of text prompts describing the objects to be detected.
For each category, we define multiple prompts that describe this class.
Afterwards, for each query prompt, a feature is computed with the text encoder.
Given the outputs $V_\sigma$ and $V_\psi$ of our model, we compute the similarity between each voxel feature with each query feature, and assign every voxel a class based on the query with the highest similarity to the voxels embedding.
We also always define a \textit{free} class that models unoccupied voxels, and set a voxel to \textit{free} when the estimated density is below a threshold $\tau$.

\section{Experiments}\label{sec:experiments}
\subsection{Dataset and Task Description}
We conduct all experiments on the nuScenes dataset \cite{caesar2020nuscenes}.
For the \textit{3D open vocabulary retrieval}, we use the benchmark provided by \cite{vobecky2024pop}.
It consists of 105 samples, each with an open vocabulary text query and corresponding binary labels for the LiDAR point cloud, with the goal of retrieving all 3D points that are described by the query.
The performance is measured by the mean-average-precision (mAP) for all points in the scene, and only for points visible in at least one camera (referred to as mAP (v)).
For \textit{zero-shot occupancy estimation}, we evaluate on the widely known Occ3D-nuScenes benchmark \cite{tian2023occ3d}, which provides semantic voxel labels for the nuScenes dataset.
We use a predefined vocabulary (see \cref{sec:vocabulary}) based on the classes given in the benchmark to assign a label to each voxel.
The performance is measured in geometric IoU and in mean-IoU over all categories in the benchmark.
For all experiments, we define \textit{Modes}, which indicate the modalities used during training.
$C$ refers to camera images, $L$ to semantically annotated LiDAR point clouds and $3D$ to (semantic) 3D voxel labels.

\subsection{Implementation Details}
For all tasks, we use the ResNet50 backbone \cite{he2016deep} and an image resolution of $256\times704$. 
The density and language heads $\Phi_\sigma$ and $\Phi_\psi$ each consist of three hidden layers with a dimension of size $256$.
We train each network with a batch size of $4$ for $18$ epochs.
We use a time horizon of $12$ (to the future and past) for temporal rendering, and generate $32,786$ rays per sample, randomly distributed over all temporal frames in the horizon.
For each ray, we sample $100$ points, and use the \textit{nerfacc} \cite{li2023nerfacc} package for rendering.
Results are provided for our model using the \textit{Full} embedding space and the \textit{Reduced} space when applying the feature subspace learning strategy.
We use the same fixed vocabulary to train the reducer $U$ for each experiment, which is based on the target classes of the Occ3D-nuScenes dataset.
The reduced dimension size is set to $L' = 128$.

\subsection{3D Open Vocabulary Retrieval}\label{sec:exp1}
\begin{figure*}
    \centering
    \includegraphics[page=4, trim=0cm 9.08cm 5.7cm 0cm, clip, width=1\textwidth]{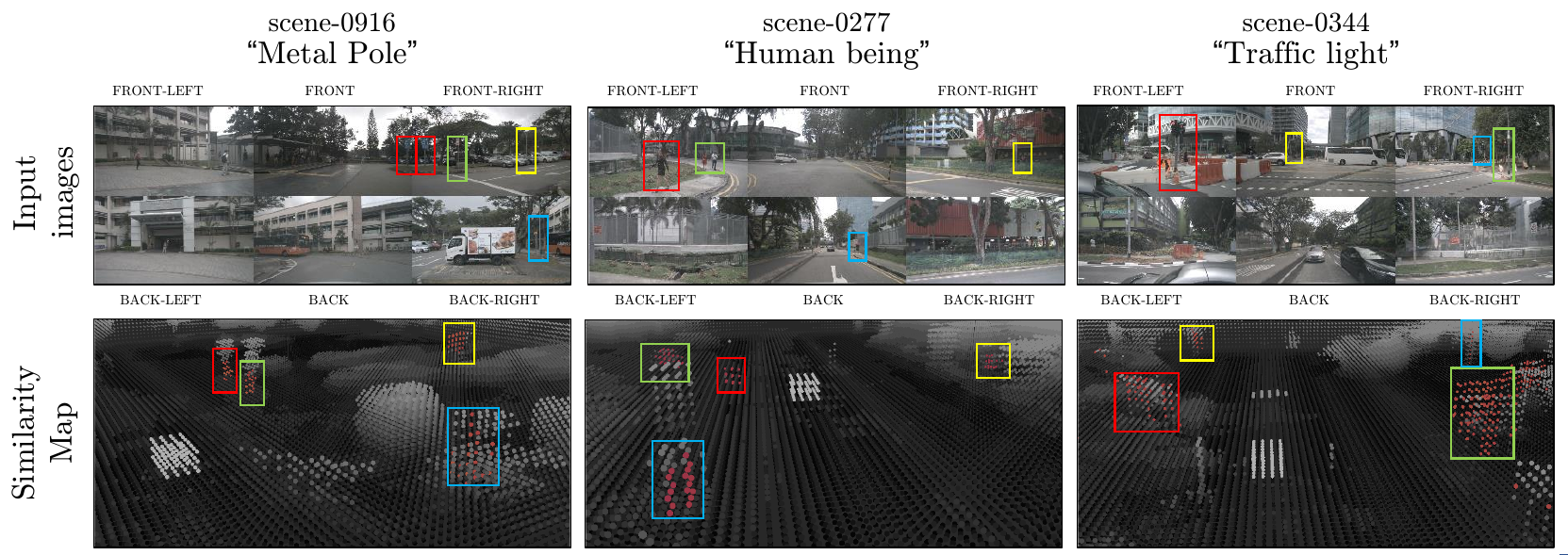}
    \caption{
        \textbf{Qualitative results showing open vocabulary retrieval on nuScenes \cite{caesar2020nuscenes}.}
        Given a text query, we compute similarities between the text embedding and each estimated voxel embedding and highlight voxels with a high similarity score. 
        Ego vehicle shown in white.
    }
    \label{fig:qual_openvoc}
\end{figure*}

We compare our results to POP-3D \cite{vobecky2024pop} on the benchmark provided by the authors. 
Their model is based on TPV-Former \cite{huang2023tri} and replaces the semantic head with a vision-language head similar to our model.
The authors train their model using LiDAR scans available in nuScenes, both for learning geometry and for the feature distillation.
They also provide results for directly using MaskCLIP as a baseline, by projecting the LiDAR sweeps on the MaskCLIP feature maps.
Results of this comparison are provided in \cref{table:openvoc}.
As is visible, our method outperforms both baselines, even though we use just vision-based supervision.
We achieve a mAP score of $21.7$ and $22.7$ (for all points and only visible points, respectively) compared to the $17.5$ and $18.4$ of POP-3D, even though we do not use LiDAR data.
These results clearly demonstrate the effectiveness of rendering supervision in distilling vision-language features into 3D.
We account this performance gain mostly to the temporal rendering approach that allows our model to learn from many overlapping views to enhance both geometry and vision-language understanding.
As expected, using the proposed feature subspace learning method decreases the open vocabulary performance of LangOcc, as we define a lower dimensional space on a specific set of classes, which decreases detection performance for open vocabulary queries that were not part of that set.
As will be shown later, the \textit{Reduced} version instead increases performance on the semantic occupancy estimation task.
We show some qualitative results in \cref{fig:qual_openvoc} highlighting the open vocabulary capabilities of LangOcc.
Given just images as input (and during training), the model estimates the 3D geometry and generic semantics around the vehicle, enabling to segment any object of interest given a text prompt.
The model keeps all the vision-language capabilities of CLIP even in 3D space, and is also capable of segmenting small and thin objects like "metal poles" accurately.
Additional qualitative results comparing CLIP features and estimated vision-language features are available in \cref{sec:app2}.

\begin{table}
\small
\begin{center}
 \caption{
    \textbf{3D open vocabulary retrieval results on the benchmark provided by \cite{vobecky2024pop}.}
    \textit{mAP (v)} is calculated only on points visible to one of the cameras.
    The \textit{Mode} indicates the modality used to train the model.
    \textit{L} and \textit{C} refer to LiDAR scans and camera images, respectively.
    }
    \label{table:openvoc}
    \begin{tabular}{lccc}
        \hline
        Method & Mode & mAP &  mAP (v) \\
        \hline
        MaskCLIP \cite{zhou2022extract} & L & - & 14.9\\
        POP-3D \cite{vobecky2024pop} & L & 17.5 & 18.4\\
        \hline
        LangOcc (Full) & C & \textbf{21.7}  & \textbf{22.7}  \\
        LangOcc (Reduced) & C & 16.6 & 18.2  \\
        \hline
    \end{tabular}
\end{center}
\vspace{-5mm}
\end{table}

\subsection{Zero-shot Semantic Occupancy Estimation} \label{sec:exp2}
We evaluate our approach against other recent approaches on the Occ3D-nuScenes dataset \cite{tian2023occ3d} and show the results in \cref{table:occ3d}.
Using vision-only training (C), our proposed method surpasses the self-supervised competitors SelfOcc \cite{huang2023selfocc} and OccNeRF \cite{zhang2023occnerf} on both geometric IoU and semantic mIoU.
LangOcc achieves a geometric IoU score of at least $51.59$, showing that our model is able to estimate the scene geometry well without any photometric losses or explicit depth supervision.
Both SelfOcc and OccNeRF explicitly supervise geometry, for example via multi-view stereo losses and RGB rendering.
Seemingly, the density learned via volume rendering of vision-language features gives sufficient signal and is even better suited than using photometric losses to learn geometry.
We hypothesize this is likely due to the high representational power of CLIP embeddings and because our model is forced to learn consistent features in 3D over many overlapping views.
As a consequence the model gets better geometry supervision compared to usually very ambiguous photometric losses.
We further provide a comparison between RGB and feature distillation losses for geometry in \cref{sec:geom}.
Furthermore, both SelfOcc and OccNeRF use class-specific segmentation networks to estimate the voxel labels, while LangOcc can theoretically detect any class with the same model (in the \textit{full} variation).
Even though our model is trained without any explicit class definition, we outperform both competitors also in terms of semantic mIoU, highlighting the power of the estimated features.
By specifying a vocabulary for the given task and using the proposed dimensionality reduction method (\cref{sec:lowerspace}), we can further increase the semantic mIoU score from $10.71$ to $11.84$.
To conclude, using just a single loss function and a straightforward training paradigm, our method achieves state-of-the-art performance on vision-only Occ3D-nuScenes, while still being capable of open-vocabulary detection.
As mentioned above, the reducer $U$ finishes training within a second and thus does not impose any notable overhead.
We also present results for methods trained with 3D voxel labels \cite{huang2021bevdet, huang2023tri, tian2023occ3d} to demonstrate the gap between supervised and self-supervised, vision-only approaches.
We provide the performance on each individual class and additional extensive qualitative results in \cref{sec:app1} and \ref{sec:app2}.

\begin{table}
\small
\begin{center}
 \caption{
        \textbf{Semantic occupancy estimation results on the Occ3D-nuScenes benchmark \cite{tian2023occ3d} in terms of geometric IoU and semantic mIoU.}
        The \textit{Mode} indicates the modality used during training. \textit{3D}, \textit{L}, \textit{C} refer to semantic 3D voxel labels, semantic LiDAR point clouds and camera images, respectively.
        Best and second best performing method per \textit{Mode} in \textbf{bold} and \textit{italics}, respectively.
        }
        \label{table:occ3d}
        \begin{tabular}{lccc}
            \hline
            Method & Mode & IoU &  mIoU \\
            \hline
            OccFormer \cite{huang2021bevdet} & 3D & - & 21.93\\
            TPVFormer \cite{huang2023tri} & 3D & - & \textit{27.83}\\
            CTF-Occ \cite{tian2023occ3d} & 3D & - & \textbf{28.53}\\
            \hline
            TPVFormer \cite{huang2023tri} & L & \textbf{17.2} & 13.57\\
            RenderOcc \cite{pan2023renderocc} & L & - & \textit{23.93}\\
            OccFlowNet \cite{boeder2024occflownet} & L & - & \textbf{26.14}\\
            \hline
            SelfOcc \cite{huang2023selfocc} & C & 45.01 & 9.30\\
            OccNeRF \cite{zhang2023occnerf} & C & - & 10.13 \\

            LangOcc (Full) & C & \textit{51.59}  & \textit{10.71} \\
            LangOcc (Reduced) & C & \textbf{51.76}  & \textbf{11.84} \\
            \hline
        \end{tabular}
    \end{center}
    \vspace{-7mm}
\end{table}

\subsection{Ablations}
\paragraph{Loss function}
We provide a comparison between using our proposed \textit{Cosine Similarity Guided MSE} function and using either the MSE loss or the Cosine Similarity loss by training a model with each loss function.
As the results in \cref{table:ablation_loss} show, our loss function leads to increased performance on each metric.

\begin{table}
\small
\begin{center}
 \caption{
        \textbf{Ablation on the loss function used for $\mathcal{L}_{lang}$.}
        }
        \label{table:ablation_loss}
        \begin{tabular}{l|ccc}
            \hline
            Loss Function & MSE & CosSim &  Cos-guided MSE \\
            \hline
            IoU & 50.29 & 49.88 & \textbf{51.59} \\
            mIoU & 9.41 & 9.89 & \textbf{10.71}\\
            mAP (v) & 20.1 & 22.6 & \textbf{22.7} \\
            \hline
        \end{tabular}
    \end{center}
\end{table}

\paragraph{Temporal Horizon}
We ablate the temporal horizon of the model during training and show the results in \cref{table:ablation_temporal}.
As expected, using no temporal rendering at all leads to very poor results, as the model can hardly learn any 3D geometry from the very few overlapping rays.
Adding 4 future and past frames during rendering supervision already improves all scores significantly, such that LangOcc achieves a better open vocabulary retrieval performance than POP-3D \cite{vobecky2024pop}.
The best performance on all tasks was achieved by using a horizon of 12, which seems to be a good trade-off between overlap of cameras and view diversity.
Adding more temporal frames led to a decrease in performance.
We hypothesize this is due to the large distance between the camera poses, so that many rays are not visible anymore in the current frame and rays intersect less overall.

\begin{table}
\small
\begin{center}
 \caption{
        \textbf{Ablation on the temporal horizon.}
        }
        \label{table:ablation_temporal}
        \resizebox{\columnwidth}{!}{
            \begin{tabular}{l|cccccc}
                \hline
                Horizon & 0 & 4 & 8 & 12 & 16 & 20 \\
                \hline
                IoU & 15.78 & 40.85 & 49.71 & \textbf{51.59} & 50.54 & 49.74 \\
                mIoU & 2.88 & 8.46 & 9.95 &  \textbf{10.71} & 9.46 & 9.16\\
                mAP (v) & 9.8 & 20.0 & 22.6 & \textbf{22.7} & 21.8 & 20.2 \\
                \hline
            \end{tabular}
        }
    \end{center}
\end{table}

\paragraph{Reduced Dimension Size}
\Cref{table:ablation_reducer} shows a comparison between using different subspace dimension sizes for the reducer $U$.
We use the same vocabulary as in \cref{sec:exp2} for all models to train the autoencoder, but modify the target dimension size (with 512 being the \textit{full} space).
As observable, the open vocabulary performance decreases steadily, the smaller the subspace gets, as the model loses representational power and overfits more on the provided vocabulary.
However, using the dimensionality reducer can offer improved performance on the zero-shot occupancy estimation task.
The best performance can be achieved at $L'=128$, which seems to be the optimal trade-off between task-specific expressiveness and not overfitting.
Decreasing the dimension size further however, up to $32$, still offers increased mIoU performance compared to the original space.
Only when the subspace dimensionality is decreased to $16$, the performance decreases drastically.
Interestingly, the geometric estimations only differ slightly from the full space at higher dimensions, while they decline at lower sizes.
This is likely because higher dimensions have more capacity to encode information about geometry, while the lower dimensions have to focus more on the semantic features from the vocabulary.
Also, feature vectors are more distinct in high dimensions, such that finding corresponding points in different views is much easier than in lower dimensions, where many feature vectors are similar.

\begin{table}
\small
\begin{center}
 \caption{
        \textbf{Ablation on the subspace dimensionality $L'$.}
        }
        \label{table:ablation_reducer}
        
            \begin{tabular}{l|cccccc}
                \hline
                L' & 16 & 32 & 64 & 128 & 256 & 512 \\
                \hline
                IoU & 50.00 & 50.18 & 51.02 & \textbf{51.76} & 51.11& 51.59 \\
                mIoU & 7.52 & 10.86 & 11.18 & \textbf{11.84} & 11.08 & 10.71 \\
                mAP (v) & 10.6 & 11.2 & 17.1& 18.2 & 19.1 & \textbf{22.7}\\
                \hline
            \end{tabular}
    \end{center}
    \vspace{-4mm}
\end{table}

\paragraph{Geometry Supervision} \label{sec:geom}
To show that training our proposed model with just the feature distillation loss $\mathcal{L}_lang$ leads to state-of-the-art 3D geometry estimations, we directly compare our approach with using photometric losses for training, like is done in \cite{huang2023selfocc, zhang2023occnerf}.
We train our proposed model with RGB rendering by replacing the language-feature head $\Phi_\psi$ with an RGB head that estimates the appearance of a voxel in terms of RGB and train with a \textit{MSE} loss on the rendered RGB values. 
As is common in NeRF approaches, we choose to model the appearance with spherical harmonics \cite{fridovich2022plenoxels,tancik2023nerfstudio}.
We compare the model on the geometric IoU score on the Occ3D-nuScenes benchmark.
Training a model with this RGB supervision leads to a geometric IoU score of $39.96$.
Our proposed supervision method leads to a significantly better IoU score of $51.59$, which confirms that the feature distillation loss provides a better supervision signal for the scene geometry than a photometric loss in our model architecture.
We speculate that this originates from the rich information of vision-language features and their independence from the viewing angle that impose clear constraints on the scene geometry.
Photometric losses on the other hand suffer from ambiguities like low-texture regions, locally similar pixel colors, different lighting conditions and dependence on the viewing angle which makes it hard to extract a clear geometric signal.

\section{Conclusion} \label{sec:conclusion}
In this paper, we have proposed a novel model that enables a generic open vocabulary scene representation and a self-supervised training mechanism that requires only images as input.
By using differentiable volume rendering, we distill the rich knowledge of the vision-language encoder CLIP into a 3D occupancy estimation model and simultaneously learn to estimate scene geometry, without any explicit geometry supervision.
This allows for generic 3D scene representations which are completely independent of specific class definitions.
Our model learns to represent the scene in a zero-shot manner, with no per-scene optimization like prior work \cite{kerr2023lerf, zhou2024feature}.
It significantly outperforms previous attempts for open vocabulary occupancy without using any LiDAR data.
Additionally, we set the new state-of-the-art performance on vision-only semantic occupancy estimation on the Occ3D-nuScenes dataset, and further improve segmentation performance using the proposed feature subspace learning method.
We conclude that by distilling knowledge of strong 2D visual encoders into 3D occupancy estimation models, stronger occupancy estimations are possible than with photometric methods like \cite{huang2023selfocc, zhang2023occnerf}.
Incorporating more generic feature representations like DINO (which has been shown to encode better geometric features than CLIP \cite{el2024probing}) can be a promising future direction.
Also, our work still lacks a mechanism to deal with dynamic objects, which leads to inconsistent supervisory signals during temporal rendering.
In future work, the explicit modeling of scene dynamics could help to remove temporally inconsistent signals, or even help estimating scene flow for a downstream planner.
Moreover, the benchmark provided by \cite{vobecky2024pop} is relatively small and covers only common driving scene objects.
To compare future open vocabulary approaches a larger and more diverse benchmark dataset would be beneficial.
Finally and building on the excellent performance of our model, additional research on open vocabulary occupancy estimation is required to further investigate its applicability and potential performance gains, highly demanded in a variety of tasks such as autonomous driving.
{
    \small
    \bibliographystyle{ieeenat_fullname}
    \bibliography{main}
}

\appendix
\renewcommand\thefigure{\thesection.\arabic{figure}}
\setcounter{figure}{0}
\renewcommand\thetable{\thesection.\arabic{table}}
\setcounter{table}{0}

\section{Details: Zero-shot Semantic Occupancy Estimation}\label{sec:app1}
We provide detailed results for each class in the Occ3D-nuScenes dataset in \cref{table:detailed}.
We outperform SelfOcc and OccNeRF in terms of geometric reconstruction and semantic segmentation performance already with the full embedding space, by just using the feature distillation loss.
When using a predefined vocabulary (cf. \cref{sec:vocabulary}) to train the proposed dimensionality reducer, we further increase the performance on this task.
We want to highlight that LangOcc can segment small, dynamic and vulnerable objects like bicylce, pedestrians and motorcylce much better than the other self-supervised methods, which is reflected in the IoU score of these classes.
These classes are usually more difficult to segment, but are also more important than background objects for downstream planners.
On the other hand, our model seems to have difficulties in segmenting the sidewalk from the driveable street accurately.
We think this comes from the fact that the text prompts we use to detect the sidewalk and the driveable surface are very similar in the CLIP space, leading to confusions.
The self-supervised methods are still far behind supervised approaches using annotated LiDAR and voxel labels in terms of performance, but are much more scalable and are independent of expensive data acquisition.

\begin{table*}
    \begin{center}
        \caption{
            \textbf{Semantic occupancy estimation performance on the Occ3D-nuScenes.}
            Performance is measured in \%IoU, best performing per column and category in \textbf{bold}, second best in \textit{italics}.}
        \label{table:detailed}
        
        \resizebox{\textwidth}{!}{%
            \addtolength{\tabcolsep}{2pt}
            \begin{tabular}{lc|cc|ccccccccccccccccc}
                \hline
                \noalign{\smallskip}
                Method & \rotatebox{90}{Mode} & IoU & mIoU & \rotatebox{90}{others} & \rotatebox{90}{barrier} & \rotatebox{90}{bicycle} & \rotatebox{90}{bus} & \rotatebox{90}{car} & \rotatebox{90}{cons. vehicle} & \rotatebox{90}{motorcycle} & \rotatebox{90}{pedestrian} & \rotatebox{90}{traffic cone} & \rotatebox{90}{trailer} & \rotatebox{90}{truck} & \rotatebox{90}{driv. surf.} & \rotatebox{90}{other flat} & \rotatebox{90}{sidewalk} & \rotatebox{90}{terrain} & \rotatebox{90}{manmade} & \rotatebox{90}{vegetation}\\
                \noalign{\smallskip}
                \hline
                \noalign{\smallskip}
                OccFormer \cite{zhang2023occformer} & 3D & - & 21.93 & 5.9 & 30.3 & 12.3 & 34.4 & 39.2 & 14.4 & 16.5 & 17.2 & 9.3 & 13.9 & 26.4 & 51.0 & 31.0 & 34.7 & 22.7 & 6.8 & 7.0 \\
                TPVFormer \cite{huang2023tri} & 3D & - & 27.83 & 7.2 & 38.9 & 13.7 & 40.8 & \textbf{45.9} & 17.2 & 20.0  & 18.8 & 14.3 & 26.7 & 34.2 & 55.6 & 35.5 & 37.6 & 30.7 & 19.4 & 16.8 \\
                CTF-Occ \cite{tian2023occ3d} & 3D & - & 28.53 &8.1 & 39.3 & 20.6 & 38.3 & 42.2 & 16.9 & 24.5 & 22.7 & 21.0 & 23.0 & 31.1 & 53.3 & 33.8 & 38.  & 33.2 & 20.8 & 18.0 \\
                \noalign{\smallskip}
                \hline
                \noalign{\smallskip}
                TPVFormer \cite{huang2023tri} & L & 17.20 & 13.57 & 0.0 & 14.8 & 9.4 & 21.3 & 16.8 & 14.5 & 13.8 & 11.2 & 5.3 & 16.1 & 19.7 & 10.8 & 9.4 & 9.5 & 11.2 & 16.5 & 17.0 \\
                RenderOcc \cite{pan2023renderocc} & L & - & 23.93 & 5.7 & 27.6 & 14.4 & 19.9 & 20.6 & 12.0 & 12.4 & 12.1 & 14.3 & 20.8 & 18.9 & 68.8 & 33.4 & 42.0 & 43.9 & 17.4 & 22.6\\
                OccFlowNet \cite{boeder2024occflownet} & L & - & 26.14 & 3.2 & 28.8 & \textit{22.2} & 28.0 & 21.7 & 17.2 & 19.6 & 11.0 & 18.0 & 24.1 & 22.0 & 67.3 & 28.7 & 40.0 & 41.0 & 26.2 & 25.6 \\
                \noalign{\smallskip}
                \hline
                \noalign{\smallskip}
                SelfOcc \cite{huang2023selfocc} & C & 45.01 & 9.30 & 0.0 & 0.2 & 0.7 & 5.5 & 12.5 & 0.0 & \textit{0.8} & 2.1 & 0.0 & 0.0 & 8.3 & \textbf{55.5} & 0.0 & \textbf{26.3} & \textit{26.6} & 14.2 & 5.6 \\
                OccNeRF \cite{zhang2023occnerf} & C & - & 10.13 & 0.0 & 0.8 &  0.8 &  5.1 & 12.5 &  \textbf{3.5} &  0.2 &  3.1 &  1.8 &  0.5 &  3.9 & \textit{52.6} & 0.0 & \textit{20.8} & 24.8 & \textit{18.4} & 13.2 \\
                LangOcc (Full) & C & \textit{51.59} & \textit{10.71} & 0.0 & \textit{2.7} & \textit{7.2} & \textit{5.8} & \textit{13.9} & \textit{0.5} & \textbf{10.8} & \textbf{6.4} & \textit{8.7} & \textit{3.2} & \textbf{11.0} & 42.1 & \textit{1.6} & 12.5 & \textbf{27.2} & 14.1 & \textit{14.5} \\
                LangOcc (Reduced) & C & \textbf{51.76} & \textbf{11.84} & 0.0 & \textbf{3.1} & \textbf{9.0} & \textbf{6.3} & \textbf{14.2} & 0.4 & \textbf{10.8} & \textit{6.2} & \textbf{9.0} & \textbf{3.8} & \textit{10.7} & 43.7 & \textbf{2.23} & 9.5 & 26.4 & \textbf{19.6} & \textbf{26.4} \\
                \noalign{\smallskip}
                \hline
            \end{tabular}
            \addtolength{\tabcolsep}{2pt}
        }
    \end{center}
\end{table*}

\section{Additional Qualitative Results}\label{sec:app2}
\begin{figure*}
    \centering
    \includegraphics[page=2, trim=0cm 9.88cm 2.41cm 0cm, clip, width=1\textwidth]{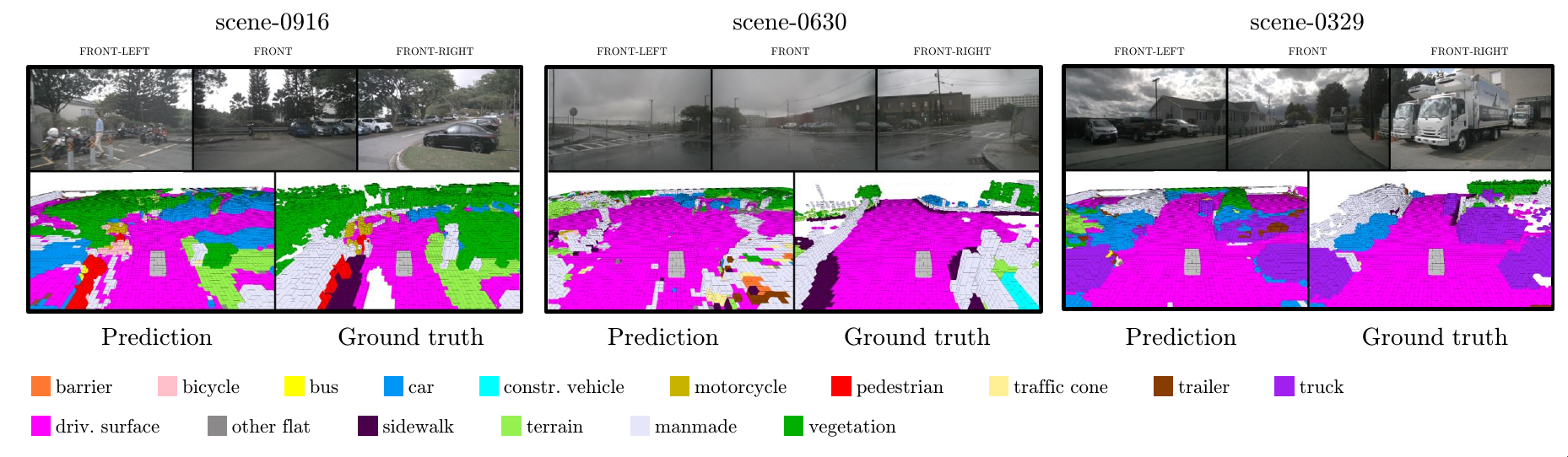}
    \caption{
        \textbf{Qualitative results showing zeros-shot semantic occupancy estimations.}
    }
    \label{fig:qual_occ3d1}
\end{figure*}

\begin{figure*}
    \centering
    \includegraphics[page=3, trim=0cm 9.16cm 11.27cm 0cm, clip, width=1\textwidth]{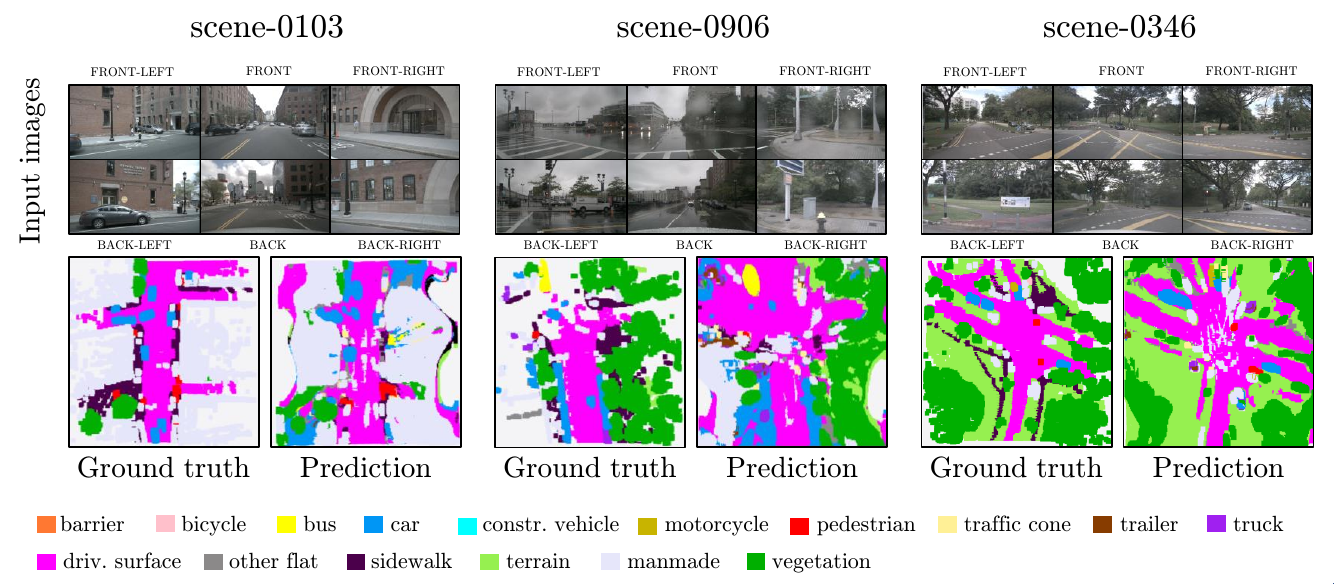}
    \caption{
        \textbf{Qualitative results showing zero-shot semantic occupancy estimations.}
    }
    \label{fig:qual_occ3d2}
\end{figure*}

\begin{figure*}
    \centering
    \includegraphics[page=5, trim=0cm 10.8cm 8.55cm 0cm, clip, width=1\textwidth]{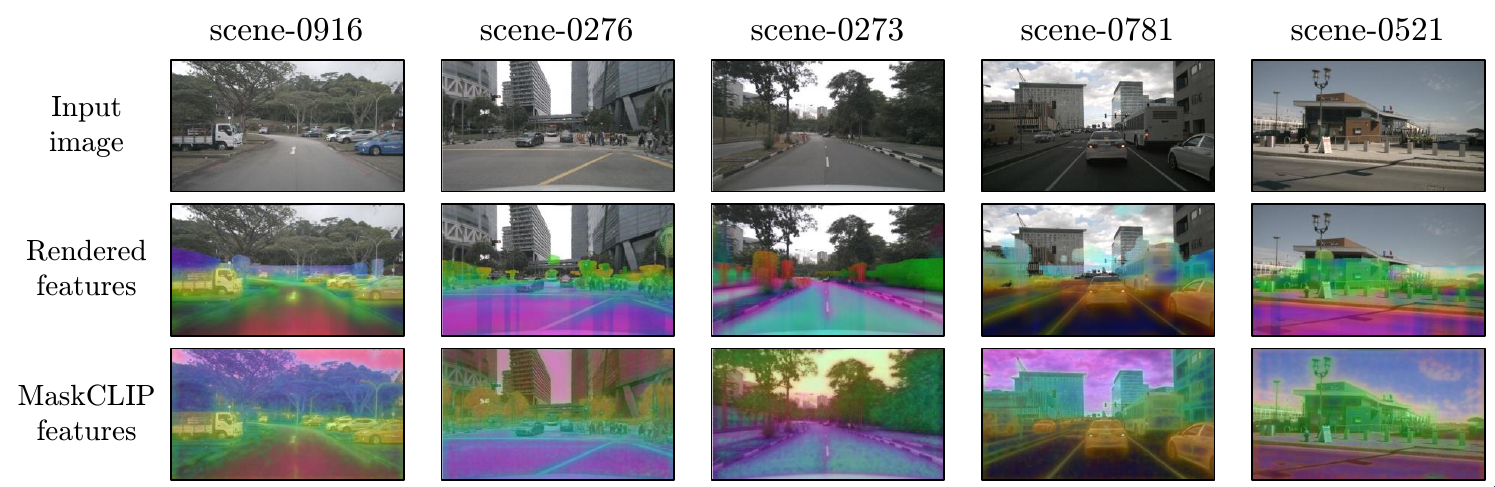}
    \caption{
        \textbf{Qualitative results depicting rendered estimated 3D features and ground truth features in 2D image space.}
        As is visible, given just the input image, our model can replicate the original CLIP embeddings accurately.
        However, our model estimates them in full 3D space.
    }
    \label{fig:qual_render}
\end{figure*}

In \cref{fig:qual_occ3d1} we present a comparison between semantic occupancy estimations of LangOcc (Reduced) and the ground truth voxel labels of Occ3D-nuScenes on different scenes, depicted from a third-person view of the ego vehicle.
\Cref{fig:qual_occ3d2} provides additional qualitative results from a birds-eye-view perspective.
Despite the model has never seen any semantic or voxel labels, and the lack of explicit supervision for depth or geometry, the model can estimate the scene geometry well, and detect most semantic features in the scenes. 
The model can even generalize to areas behind occluding objects to a certain extend.
However, it is clearly visible that the semantic predictions are still fairly noisy, which likely is a result of the sometimes ambiguous vision-language features.
\Cref{fig:qual_render} illustrates feature maps of our model when rendering the estimated voxel features back to the image space, in comparison to the ground truth feature maps extracted via MaskCLIP.
The features are reduced to three dimensions using PCA for visualisation (on each column individually).
Note that the rendered feature maps are not the output of the model, but LangOcc outputs vision-language features in 3D voxel space.
The estimated feature maps are generated via volume rendering our predictions into the input cameras.
These feature maps are essentially the input into the loss function, which can be backpropagated through the whole model, as the volume rendering is fully differentiable.
One can see that the rendered feature maps retain the expressiveness of the original vision-language aligned feature maps, even in 3D space.
We therefore inherit all the vision-language capabilities of the original feature space, enabling \textit{open vocabulary occupancy}.


\section{Vocabulary} \label{sec:vocabulary}
We present the vocabulary that we use for semantic occupancy estimation on Occ3D-nuScenes and to train the \textit{Reducer} in \cref{table:vocabulary}.
For each class in the Occ3D-nuScenes benchmark, we define a set of text prompts that describe that category.
As described in the paper, during inference, we compare the estimated voxel features with each text embedding of the vocabulary, and assign each voxel the label belonging to the prompt with the highest similarity score.
To train the \textit{Reducer}, we concatenate all text prompts to form a single dataset of text embeddings, on which the \textit{Reducer} is trained.
\begin{table*}
\small
\begin{center}
 \caption{
    \textbf{Vocabulary used for zero-shot semantic occupancy estimation and to train the \textit{Reducer}.}}
    \label{table:vocabulary}
    
        \begin{tabular}{l|l}
            \hline
            Class & Prompts \\
            \hline
            'car' & \makecell[ll]{'Vehicle designed primarily for personal use.', 'car', 'vehicle', 'sedan', 'hatch-back', 'wagon',\\ 'van', 'mini-van', 'SUV', 'jeep'}  \\
            \hline
            'truck' & 'Vehicle primarily designed to haul cargo.', 'pick-up', 'lorry', 'truck', 'semi-tractor' \\
            \hline
            'trailer' &  'trailer', 'truck trailer', 'car trailer', 'bike trailer' \\
            \hline
            'bus' & 'Rigid bus', 'Bendy bus' \\
            \hline
            'construction\_vehicle' & 'Vehicle designed for construction.', 'crane' \\
            \hline
            'bicycle' & 'Bicycle' \\
            \hline
            'motorcycle' & 'motorcycle', 'vespa', 'scooter' \\
            \hline
            'pedestrian' & 'Adult.', 'Child.', 'Construction worker', 'Police officer.'\\
            \hline
            'traffic\_cone' & 'traffic cone.' \\
            \hline
            'barrier' & 'Temporary road barrier to redirect traffic.', 'concrete barrier', 'metal barrier', 'water barrier' \\
            \hline
            'driveable\_surface' & 'Paved surface that a car can drive.', 'Unpaved surface that a car can drive.' \\
            \hline
            'other\_flat' & 'traffic island', 'delimiter', 'rail track', 'small stairs', 'lake', 'river' \\
            \hline
            'sidewalk' & 'sidewalk', 'pedestrian walkway', 'bike path' \\
            \hline
            'terrain' & 'grass', 'rolling hill', 'soil', 'sand', 'gravel' \\
            \hline
            'manmade' & \makecell[ll]{'man-made structure', 'building', 'wall', 'guard rail', 'fence', 'pole', 'drainage', 'hydrant','flag', \\ 'banner', 'street sign', 'electric circuit box', 'traffic light', 'parking meter','stairs'}  \\
            \hline
            'vegetation' & 'bushes', 'bush', 'plants', 'plant', 'potted plant', 'tree', 'trees'\\
            \hline
            'background' & \makecell[ll]{'Any lidar return that does not correspond to a physical object, such as dust, vapor, noise, fog, raindrops, \\ smoke and reflections.','sky'}\\
            \hline
        \end{tabular}
    \end{center}
\end{table*}

\end{document}